\pdfoutput=1
\documentclass[12pt]{article}

\usepackage{acl}
\usepackage{latexsym, graphicx, amsmath, float, tabularx, subcaption}
\usepackage{booktabs} % To thicken table lines
\usepackage{threeparttable} % to use table notes
\usepackage{times}
\usepackage{latexsym}

\usepackage[T1]{fontenc}
\usepackage[utf8]{inputenc}

\usepackage{microtype}

\usepackage{inconsolata}

\title{SLIDE: A Framework Integrating Small and Large Language Models for Open-Domain Dialogues Evaluation}

\newcommand*\samethanks[1][\value{footnote}]{\footnotemark[#1]}

\author{ 
    Kun Zhao\textsuperscript{1}\thanks{\quad \small Equal contribution}\space,
    Bohao Yang\textsuperscript{2}\samethanks\space,
    Chen Tang\textsuperscript{2},
    Chenghua Lin\textsuperscript{2}\thanks{\quad \small Corresponding authors}\;,
     \textbf{Liang Zhan}\textsuperscript{\textbf{1}}\samethanks\space\\ 
\textsuperscript{1} Department of Electrical and Computer Engineering, University of Pittsburgh, US\vspace{-0.5mm} \\ 
\textsuperscript{2} Department of Computer Science, The University of Manchester, UK\vspace{-0.5mm} \\
\texttt{
\{kun.zhao, liang.zhan\}@pitt.edu,  
}\vspace{-0.5mm} 
\texttt{
bohao.yang-2@postgrad.manchester.ac.uk
}\vspace{-0.5mm}\\
\texttt{
\{chen.tang, chenghua.lin\}@manchester.ac.uk
}\vspace{0.4mm} 
}

\begin{document}
\maketitle
\begin{abstract}

The long-standing one-to-many problem of gold standard responses in open-domain dialogue systems presents challenges for automatic evaluation metrics. 
Though prior works have demonstrated some success by applying powerful Large Language Models (LLMs), existing approaches still struggle with the one-to-many problem, and exhibit subpar performance in domain-specific scenarios. We assume the commonsense reasoning biases within LLMs may hinder their performance in domain-specific evaluations. To address both issues, we propose a novel framework \textbf{SLIDE} (\textbf{S}mall and \textbf{L}arge \textbf{I}ntegrated for \textbf{D}ialogue \textbf{E}valuation), that leverages both a small, specialised model (SLM), and LLMs for the evaluation of open domain dialogues. 
Our approach introduces several techniques: (1) Contrastive learning to differentiate between robust and non-robust response embeddings; (2) A novel metric for semantic sensitivity that combines embedding cosine distances with similarity learned through neural networks, and (3) A strategy for incorporating the evaluation results from both the SLM and LLMs. 
Our empirical results demonstrate that our approach achieves state-of-the-art performance in both the classification and evaluation tasks, and additionally the SLIDE evaluator exhibits better correlation with human judgements. Our code is available at \textbf{\url{https://github.com/hegehongcha/SLIDE-ACL2024}}.

% The final score, named SLIDE, is derived by merging these two scores. Our experimental results demonstrate that our SLM not only surpasses state-of-the-art performance in classification tasks on the DailyDialog++ dataset, but also that the SLIDE evaluator achieves superior correlation with human assessments.

\end{abstract}

% =============================== Section 1 ==================================
\section{Introduction}

Open-domain dialogue generation is an important research topic in the field of Natural Language Processing (NLP)~\cite{zeng-etal-2021-affective, wang-etal-2021-fast, xiao2023evaluating, tang2023enhancing,  tang2023improving, yang2024improving}. Evaluating such dialogues, however, is challenging due to the prevalent one-to-many issue where one conversational context may have multiple reasonable, yet semantically different responses. Additionally, adversarial negative responses, which have word overlap with given contexts, also raise further challenges in adequately evaluating open-domain dialogue~\cite{Sai2020ImprovingDE}.
% Existing traditional methods can be divided into two categories: reference-free and reference-based. Reference-based methods include approaches centred on word-overlap (e.g., BLEU \cite{papineni-etal-2002-bleu}, ROUGE \cite{lin-2004-rouge} and METEOR \cite{banerjee-lavie-2005-meteor}) and semantic embeddings (e.g., BERTScore \cite{Zhang2020BERTScoreET} and BARTScore \cite{yuan2021bartscore}). 

Existing evaluation methods centered on word-overlap, such as BLEU \cite{papineni-etal-2002-bleu}, ROUGE \cite{lin-2004-rouge}, METEOR \cite{banerjee-lavie-2005-meteor}) and semantic-embedding based approaches (e.g., BERTScore \cite{Zhang2020BERTScoreET} and BARTScore \cite{yuan2021bartscore}), evaluate responses via calculating the similarity between the generated response and contexts or gold references. Therefore, they struggle to adequately address the one-to-many problem as a context may permit many different responses. 
% On the other hand, reference-free methods evaluate dialogues by calculating the relationship between context and response pairs, and include metrics such as RUBER \cite{Tao2018}, MAUDE \cite{sinha-etal-2020-learning}, and EMS \cite{chan-etal-2021-enhancing}. These methods are trained on single-response datasets, leading to difficulty in handling multiple golden references.
% Recently, many novel evaluation approaches have been presented that utilise the unique capabilities of Large Language Models (LLM)~\cite{Fu2023GPTScoreEA, Liu2023GEvalNE}.
% In recent years, a range of new studies have emerged, seeking to utilise the commonsense reasoning capabilities learned in Large Language Models (LLM) for automatic evaluation~\cite{Fu2023GPTScoreEA, Liu2023GEvalNE}.
Recently, there are some novel approaches proposed to leverage the commonsense reasoning capabilities learned in Large Language Models (LLMs) for a wide range of NLP tasks, such as dialogue generation, evaluation and sentiment analysis~\cite{loakman2023iron, liu-etal-2023-g, yang-etal-2024-effective-distillation, Chiang2023CanLL, yang2024structured}.
% \citet{Fu2023GPTScoreEA} proposed GPTScore, which based on a large pre-trained language model, achieve multi-aspect, customised, and training-free evaluation.
% \citet{Wang2023IsCA} conducted a preliminary study to verify the effectiveness of LLM-based evaluator.
% \citet{Kocmi2023LargeLM} use GPT models to
% evaluate machine translation. \citet{Liu2023GEvalNE} proposed GPT-Eval, an evaluator that employs \texttt{GPT-4}
% on a range of tasks, including Dialogue Response
% Generation, Text Summarization, Data-to-text generation, and Machine Translation.
% % \citet{Wang2023LargeLM} found LLM have a system bias 
% %need to rewrite
However, there are still some limitations of LLM-based evaluation. For example, \citet{Lin2023LLMEvalUM} propose LLM-EVAL to evaluate open-domain dialogue systems with a range of LLMs and identify numerous issues including the significant reliance on prompt phrasing, which may artificially limit the effectiveness of the metrics and the evaluation quality changed with prompt formulation, which makes scoring unreliable and unrobust. Additionally, \citet{Wang2023LargeLM} state that without additional support, LLMs alone do not constitute equitable evaluators.
% (2) Significant dependence on the quality and clarity of prompts and evaluation schema to facilitate effective evaluation. \citet{Wang2023LargeLM} also found LLMs have a positional bias, often leading to further unfair evaluation.
% Due to previous findings that LLM cannot effectively resolve the one-to-many problem alone,
% In addition, we found LLM struggles to generate a precise and reliable evaluation when we employ them only to evaluate positive and adversarial negative responses. 

% Furthermore, it has been observed that LLMs encounter difficulties in generating precise and reliable evaluations when assessing positive and adversarial negative responses. Whilst LLMs are able to classify and evaluate adversarial negative responses, they are frequently unable to make correct predictions when encountering different positive responses given a conversational context. We also found small task-specific model (SLM), which are designed to resolve specific tasks and offer cost-effective solutions, inclined to recognise positive responses in the identical dataset after training. However, task-specific models require a large amount of data to train, and problems quickly arise regarding data scarcity and the paucity of examples for many aspects of open-domain dialogue.
In addition, LLMs face significant challenges in generating accurate and dependable assessments when evaluating both positive and adversarial negative responses. 
Although LLMs can classify and evaluate adversarial negative responses with a degree of proficiency, their performance significantly diminishes when dealing with multiple positive responses, which is the one-to-many nature of open-domain dialogue, a reasonable response that matches well to context could have significantly different semantics to its reference. 
As for small task-specific model (SLM), which is cost-effective alternatives, reveal a propensity towards positive responses. 
Nonetheless, the deployment of SLM is challenging, as they require extensive training data. This requirement often leads to issues related to data scarcity and a lack of sufficient examples covering a wide range of topics within open-domain dialogue, presenting obstacles to their effective implementation and scalability.
Therefore, we combine a SLM with LLM to solve the one-to-many issue, and we also resolve the data scarcity problem through data augmentation via LLMs.
% faced by solely using either type of model. 
Initially, we train a SLM using contrastive learning to minimise the cosine distance between the positive responses and the context embeddings, and enlarge the distance between the embeddings of adversarial negative responses and the context embeddings. After fine-tuning with contrastive learning, these compact models can discern between positive and adversarial negative responses adequately.
% Moreover, our findings indicate that positive responses exhibit a smaller cosine distance relative to the context when compared with adversarial negative responses. 
Subsequently, we divide the representations of the responses into two categories: robust and non-robust vectors. Through contrastive learning,
 % we classify these vectors, 
we only retain robust vectors for subsequent steps while excluding non-robust ones. Following this, we proceed to calculate the contextual-response distance, and obtain a probability value from the classification model. 
Finally, an integrated score is composed  by combining these two values as the final evaluation score of the SLM model. Ultimately, the evaluation score produced by the SLM and the score from the LLM is further integrated as the final evaluation score, SLIDE, used for evaluation. 

% In order to verify the superiority of our method, we design two experiments including a classification task and an evaluation task.
A series of experiments are conducted to analyse the effectiveness of our proposed framework, including a classification task and an evaluation task. For the classification task, we use the SLM to classify the positive and adversarial negative responses from the DailyDialog++ dataset. Experimental results demonstrate that our SLM achieves state-of-the-art (SOTA) performance, even when compared to the results of some LLMs (e.g., \texttt{GPT-3.5}). We additionally note that the SLM achieves higher accuracy in positive samples, whilst the LLM performs better for negative example responses. For the evaluation task, SLIDE also shows a good correlation with human scores.

Our contributions can be summarised as the following:
\begin{itemize}
    \item We propose a new evaluator for open-domain evaluation, namely \textbf{SLIDE} (\textbf{S}mall and \textbf{L}arge \textbf{I}ntegrated for \textbf{D}ialogue \textbf{E}valuation). To the best of our knowledge, this is the first attempt to combine these models in the open-domain dialogue evaluation task.
    % \item We design a new classifier to categorise dialogue responses as positive or negative responses, which integrates embedding distances and probability values to achieve state-of-the-art performance on multi-reference datasets.
    \item To better measure the semantic differences between dialogues, we introduce a novel evaluation score, which is integrated from the derivative of embedding cosine distances, and the similarity probability created by neural networks. The score has been demonstrated to be very effective in classifying positive and negative responses, leading to a SOTA result on the open-domain dialogue datasets.
    \item We augment existing dialogue evaluation datasets with multiple positive and adversarial negative responses. This enhanced dataset can be used in fine-tuning the open-domain SLM model.
    \item We conduct a range of experiments to demonstrate the proposed SLIDE effectively addresses the shortcomings of singular SLM-based or LLM-based open-domain dialogue evaluation models regarding the one-to-many reference problem.
\end{itemize}

% discard complex prompt designing about LLM-based evaluation metrics and we only , which have promoted the effectiveness of LLM.

% 4. Because of the complexity and diversity of prompts when using LLM to do evaluation, instead of using LLM to do evaluation task directly, we design a simple prompt to let LLM only focus on generation task and let small model to complete evaluation task. Our method cleverly avoids the difficulties of LLM in the evaluation task, and at the same time fully and more effectively applies LLM to the automatic evaluation metrics of open-domain dialogue. 

\section{Related Work}

%\noindent\textbf{Dialogue evaluation metrics.}~~
\subsection{Dialogue evaluation metrics}
Traditional $n$-gram-based evaluation metrics like BLEU~\cite{papineni-etal-2002-bleu}, ROUGE~\cite{lin-2004-rouge}, and METEOR~\cite{banerjee-lavie-2005-meteor} assess the overlaps of words between candidate responses and a reference standard. On the other hand, metrics based on embeddings, such as Extrema~\cite{forgues2014bootstrapping} and BERTScore~\cite{Zhang2020BERTScoreET}, transform the responses and references into high-dimensional representations to evaluate semantic similarities. 
% Despite these methods, their effectiveness in evaluating open-domain dialogues remains limited.

Recent years have witness an increasing interest in developing the trainable metrics. RUBER~\cite{Tao2018} evaluates the similarity between the generated response, its context, and the actual reference. DEB~\cite{Sai2020ImprovingDE}, a BERT-based model pre-trained on extensive Reddit conversation datasets, was introduced to improve evaluation effectiveness. The Mask-and-fill approach~\cite{Gupta2021SynthesizingAN}, utilising Speaker Aware (SA)-BERT~\cite{Gu2020SpeakerAwareBF}, aims to better understand dialogues. MDD-Eval~\cite{Zhang2021MDDEvalSO} was developed for evaluating dialogue systems across various domains by using a teacher model for annotating dialogues from different domains, thus supporting the training of a domain-agnostic evaluator. This method, however, relies on human labels and extra training data, which our proposed approach does not require. Lastly, CMN~\cite{zhao2023evaluating} was designed to effectively address the one-to-many nature of open-domain dialogue evaluation in the latent space, yet it does not account for adversarial negative examples.

\subsection{LLM-based Evaluators}
The application of LLM for evaluation purposes has seen significant interest due to its impressive capabilities across various tasks. GPTScore was introduced by \citet{Fu2023GPTScoreEA} as a method for multi-faceted, tailor-made, and training-independent evaluation. Preliminary investigations by \citet{wang-etal-2023-chatgpt} into the efficacy of evaluators based on LLM have been conducted. \citet{Kocmi2023LargeLM} have utilised GPT models for assessing machine translation quality. Furthermore, \citet{liu-etal-2023-g} developed G-EVAL, leveraging \texttt{GPT-4} to evaluate across multiple tasks such as dialogue response generation, text summarization, data-to-text generation, and machine translation. Despite these advancements, the use of LLM-based metrics to assess adversarial negative responses within open-domain dialogue scenarios remains unexplored. 
% several works have employed Large Language Models (LLM) to conduct evaluation due to their promising performance in numerous tasks. \citet{Fu2023GPTScoreEA} proposed GPTScore to achieve multi-aspect, customised, and training-free evaluation. \citet{wang-etal-2023-chatgpt} conducted a preliminary study to verify the effectiveness of an LLM-based evaluator. \citet{Kocmi2023LargeLM} use GPT models to evaluate machine translation. \citet{Liu2023GEvalNE} proposed GPT-Eval, an evaluator that employs \texttt{GPT-4} on a range of tasks, including Dialogue Response Generation, Text Summarization, Data-to-text generation, and Machine Translation. However, these LLM-based metrics have not been used to evaluate adversarial negative responses in the context of open-domain dialogue.

\section{Methodology}
\subsection{Model Architecture}
\begin{figure*}[ht]
\small
\centering 
\includegraphics[scale=0.49]{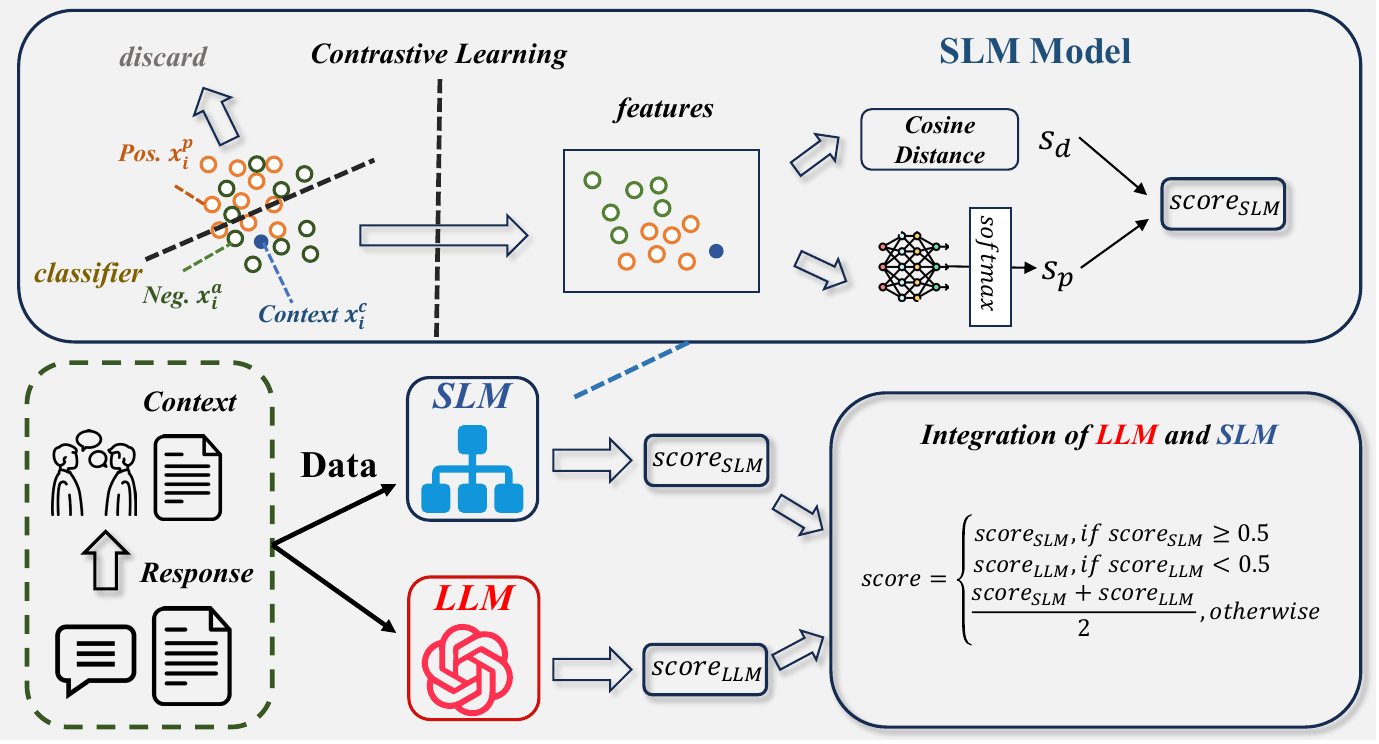}
\caption{The architecture of the proposed model. We first use an SLM trained by constrastive learning to calculate the distance between context and responses. Following this, we calculate the probability of a response being positive and the cosine distance between context and reponse, in which case we then use them to acquire $score_{SLM}$. Secondly, we use an LLM to acquire $score_{LLM}$. Finally, we acquire the final score in accordance to our findings that LLM are more inclined to recognise negative responses correctly whilst SLM recognise positive responses better. }
\label{fig:model}
\end{figure*}

As shown in ~\autoref{fig:model}, our approach can be divided into two phases: the training process and the evaluation process. During the training phase, our focus is on training the SLM to classify positive and adversarial negative responses. In the evaluation phase, we integrate the SLM and LLMs to evaluate open-domain  dialogue responses. 

\subsection{Training Process}
we first use contrastive learning to train the SLM to discern positive responses from adversarial negative responses.

% We first use contrastive learning to train an encoder.
We employ Sentence-Transformer~\cite{reimers2019sentence} to encode the context and responses. 
Specifically, we encourage the embeddings of positive responses to be closer to context embeddings, whilst negative adversarial responses move further away. Given a triplet of $<\mathbf{x}_i^c, \mathbf{x}_i^p, \mathbf{x}_i^a>, $ where  $1 \leq i \leq n$, $\mathbf{x}_i^c$ represents context, $\mathbf{x}_i^p$ represents positive responses, and $\mathbf{x}_i^a$ represents adversarial negative responses, the goal of the training process is to minimise the distance between $\mathbf{x}_i^p$ and $\mathbf{x}_i^c$ while maximising the separation between $\mathbf{x}_i^a$ and $\mathbf{x}_i^c$. The loss function is defined as follows:
\begin{align} 
    \mathbf{h}_i^{\mathrm{type}} &= \mathrm{Encoder}(\mathbf{x}_i^{\mathrm{type}}) \nonumber\\ %\indent 
    % h_i^p &= \mathrm{Encoder}(\mathbf{x}_i^p) \nonumber\\ %\indent 
    % h_i^a &= \mathrm{Encoder}(\mathbf{x}_i^a) \nonumber\\ %\indent 
    \mathcal{L}_{out} &= \mathrm{max}(||\mathbf{h}_i^c - \mathbf{h}_i^p||   \nonumber\\ %\indent 
    &- ||\mathbf{h}_i^c - \mathbf{h}_i^a|| + \mathrm{margin}, 0)
\end{align}
in which $\mathrm{margin}$ is a hyperparameter and $\mathrm{type}=\{c, p, a\}$. $\mathbf{h}_i$ is the hidden state of the $\mathbf{x}_i$, whilst $||\mathbf{h}_i^c - \mathbf{h}_i^p|| - ||\mathbf{h}_i^c - \mathbf{h}_i^a||$ refers to the cosine distance.

To enhance the precision of classification, we undertake a disentanglement process. The response embeddings are disentangled into two different sub-representations: robust embeddings and non-robust embeddings. 
% Assuming that response representations can be divided into two different sub-representations,
% the first we refer to as the robust embedding and second is the non-robust embedding. 
The robust embedding can be interpreted as a salient feature for classification, whilst the non-robust embedding constitutes noise that could act as an interfering element, potentially leading to wrong predictions.

We denote the robust and non-robust embeddings of positive responses $\mathbf{h}_i^p$ as $\{\mathbf{h}_i^{pr}, \mathbf{h}_i^{pn}\}$, respectively. On the other hand, $\{\mathbf{h}_i^{ar}, \mathbf{h}_i^{an}\}$ represents the robust and non-robust embeddings of adversarial negative responses $\mathbf{h}_i^a$. It is imperative that the robust and non-robust embeddings within both  positive responses and adversarial negative responses maintain clear distinctions.
Similarly, the separation between the robust embeddings of different types of responses should be maintained. 
 % Additionally, we also use contrastive loss in this step. According to this, 
Therefore, we define our training loss function as follows:
\begin{align}
\small
        &\mathcal{L}_{\mathrm{ins\_same\_pos}} = z_1*d_1^2 + (1-z_1)\nonumber\\ %\indent
        &*\mathrm{max}(\mathrm{margin}-d_1,0)^2 \nonumber\\ %\indent 
        &\mathcal{L}_{\mathrm{ins\_same\_neg}} = z_2*d_2^2 + (1-z_2)\nonumber\\ %\indent
        &*\mathrm{max}(\mathrm{margin}-d_2,0)^2 \nonumber\\ %\indent
        &\mathcal{L}_{\mathrm{out\_robust}} = z_3*d_3^2 + (1-z_3)\nonumber\\ %\indent
        &*\mathrm{max}(\mathrm{margin}-d_3,0)^2 
\end{align}
where $d_j, 1 \leq j \leq 3$ represents the cosine similarity and $z_j, 1 \leq j \leq 3$ indicates whether a pair of vectors match; with $z_j=0$ denoting no match and consequently a greater distance between the vectors. Conversely, $z_j=1$ signifies a match leading to a closer distance. To encourage divergence between the robust and non-robust vectors within both positive or negative classes, we set $z_1=z_2=0$. Similarly, for the robust vectors across positive and negative classes to diverge, we assign $z_3=0$.
% For the robust vectors and non-robust vectors inside positive or negative, we want the distance between them to become farther, so $z_1=z_2=0$. For the robust vectors of positive and negative, we also want the distance between them to become farther. , so $z_3=0$.
\begin{align}
    d_1 &= ||\mathbf{h}_i^{pr}-\mathbf{h}_i^{pn}|| \nonumber\\ %\indent
    d_2 &= ||\mathbf{h}_i^{ar}-\mathbf{h}_i^{an}|| \nonumber\\ %\indent
    d_3 &= ||\mathbf{h}_i^{pr}-\mathbf{h}_i^{ar}|| 
\end{align}
% We use robust embedding to calculate the distance between the context and response pairs to acquire the probability value from a classification model. 

In addition, we develop a classification network to classify each factor. The process is defined as follows:
\begin{align}
    \mathbf{h} &= \mathrm{concat}(\mathbf{h}_c, \mathbf{h}_{\mathrm{res}}) \nonumber\\ %\indent
    p_i &= \mathrm{Softmax}(\mathrm{Linear}(\mathbf{h})) \nonumber\\ %\indent
    \mathcal{L}_{\text{cls}} &= \sum_{i=0}^2 y_i * p_i
\end{align}
where $\mathbf{h}_{res}$ encompasses $\{\mathbf{h}_i^{pr}, \mathbf{h}_i^{pn}, \mathbf{h}_i^{ar}, \mathbf{h}_i^{an}\}$, while $y_i=\{0,1,2\}$ is the label.
Specifically, $y_i=1$ corresponds to $\mathbf{h}_i^{pr}$, $y_i=0$ relates to $h_i^{ar}$; for all other cases, $y_i=2$.

% For $h_i^{pr}$, $y_i=1$, for $h_i^{ar}$, $y_i=0$, otherwise $y_i=2$.
In summary, the total loss function is
\begin{align}
    \mathcal{L} &= \mathcal{L}_{\mathrm{out}} + \mathcal{L}_{\mathrm{ins\_same\_pos}} + \mathcal{L}_{\mathrm{ins\_same\_neg}} \nonumber\\ %\indent
    &+  \mathcal{L}_{\mathrm{out\_robust}} + \mathcal{L}_{\mathrm{cls}}
\end{align}

\subsection{Evaluation Process}
After training the SLM, we proceed to the evaluation phase. 
% Given a context $x_c$ and candidate response $x_r$, 
We first follow G-EVAL~\cite{liu-etal-2023-g} and prompt LLM to evaluate the given context and response. The prompt could be found in ~\ref{prompt:eva}.
Therefore we obtain a score referred to as $\mathrm{Score_{LLM}}$. Subsequently, SLM is employed to encode both the context and the response. 
This step involves segregating the response encoding into two distinct embeddings—robust and non-robust. The mathematical representation of this process is given by
% further disentangling the response embedding into a robust and non-robust embedding.
\begin{align}
    \mathbf{h}_c &= \mathrm{Encoder}(\mathbf{x}_c) \nonumber\\ %\indent
    \mathbf{h}_r &= \mathrm{Encoder}(\mathbf{x}_r) \nonumber\\ %\indent
    \mathbf{h}_{r,\mathrm{robust}}, &\mathbf{h}_{r,\mathrm{non}} = \mathrm{sep}(\mathbf{h}_r)
\end{align}

Then we calculate the cosine distance between context and response, as well as the probability for the given response.
% We then compute the cosine distance to measure the distance between the context embeddings and those of generated positive responses, as well as the distance between the context and response.
\begin{align}
    d &= \mathrm{cosine\_similarity}(\mathbf{h}_c, \mathbf{h}_{r,\mathrm{robust}}) \nonumber\\ %\indent
    s_d &= (d-d_{\mathrm{min}})/(d_{\mathrm{max}}-d_{\mathrm{min}}) \nonumber\\ %\indent
    s_p &= p(y=1) \nonumber\\ %\indent 
    &= \mathrm{Softmax}(\mathrm{Linear}(\mathbf{h}_c, \mathbf{h}_{r,\mathrm{robust}})) 
\end{align}
where $d$ is the cosine similarity between a context and a response. The normalised distance is denoted by $s_d$,  while $s_p$ represents the prediction probability of the classifier. 
% predict the label of the response for a given context. 
% Due to the training process, we know that positive responses are closer in embedding space to the context while negative responses are further away, so $d_{pos}<d_{neg}$. 
Experimental evidence from training suggests that positive responses tend to be closer in embedding space to their corresponding contexts than negative ones, implying that $d_{\mathrm{pos}}<d_{\mathrm{neg}}$. Furthermore, it has been observed that for negative responses, $s_p$ is typically lower than for positive ones. 
Consequently, it can be deduced that for positive responses, $s_d - s_p$ will be smaller than for negative ones. Based on these insights, we define a new probabilistic score for our SLM as follows:
% We also know that the $s_p$ of the negative response is less than the $s_p$ of the positive response. 
% As a result, we can infer that $s_d-s_p$ of the positive response is less than the value of the negative response. From this, we design a new probability as our SLM score:
\begin{align}
\mathrm{Score_{SLM}} = 1 - s_d + s_p
\end{align}
Finally we design an integration strategy to integrate  $\mathrm{Score_{SLM}}$ and $\mathrm{Score_{LLM}}$. Our empirical findings suggest that while the SLM demonstrates greater accuracy in identifying positive responses, the LLMs shows a greater performance on classifying negative ones. 
% Henceforth, we establish our final score utilizing this dichotomy.
% Specifically, from our experimentation, we find that the SLM is more accurate for positive responses and the LLM is more inclined to accurately discern negative responses. 
We define the final score as follows:
\begin{align}
\small
    \mathrm{Score}=\left\{\begin{array}{l}
    \mathrm{Score_{SLM}},\mathrm{Score_{SLM}} \geq 0.5, \\ %\indent
    \mathrm{Score_{LLM}},\mathrm{Score_{LLM}}<0.5,  \\ %\indent
    (\mathrm{Score_{SLM}}+\mathrm{Score_{LLM}})/2,\mathrm{otherwise}
    \end{array}\right.
\end{align}

\section{Experimental Setup}

\subsection{Dataset}
We conduct dialogue evaluation on three open-domain dialogue datasets: DailyDialog++~\cite{Sai2020ImprovingDE}, TopicalChat~\cite{Gopalakrishnan2019TopicalChatTK}, and Personachat~\cite{zhang-etal-2018-personalizing}. 
The DailyDialog++ dataset contains 9,259 contexts in the training set, 1,028 in the validation set, and 1,142 in the test set. Each context is accompanied by five positive responses, five random negative responses, and five adversarial negative responses. 
However, the PersonaChat and TopicalChat datasets lack multiple positive and adversarial negative responses. To address this issue, we employ \texttt{GPT-4} to generate both positive and adversarial negative responses and positive responses. The prompt for generating responses could be found in ~\ref{prompt:generation}. As a results, the generated training and test sets for both PersonaChat and TopicalChat datasets comprises 2,000 contexts. 
Each context within these datasets is further enriched with a set of responses, consisting of five positive responses and five adversarial negative responses. 

% Consequently, we first use LLM to generate these response types via prompting. After generation, we examine the generated responses and perform evaluation on these datasets. We then calculate both the SLM and LLM score and, in accordance with findings in the classification task, design a new strategy to get the final score: When the SLM recognises the response as a positive response, we use the SLM score as the final score; When the LLM discerns the response as a negative one, we use LLM score; Otherwise, we use the mean value of these two scores as the final score. 
% }

Validation of the generated responses is conducted through a structured three-step process. Initially, a classification prompt is designed for the LLMs to determine the validity of the positive responses generated by \texttt{GPT-4}.
Invalid responses are discarded, and we prompt the \texttt{GPT-4} for subsequent generation attempts.
% is tasked anew with generating alternative examples.
Secondly, we randomly select 1,200 generated responses from the test set for assessment by human annotators. A significant majority of these responses are appropriate to their contexts,  exceeded 98\%. The final phase involves a comparative analysis with the human-annotated benchmark dataset DailyDialog++. 
A SLM trained on training set of the DailyDialog++ dataset with 88\% accuracy is utilised to classify the generated responses without incorporating distance metrics. The classification accuracy is 83\% on the PersonaChat test set and 85\% on the TopicalChat test set. 
% The accuracy closely mirroring the 88\% accuracy achieved on the DailyDialog++ test set. 

% Additionally, we calculate BLEU-1 score to measure  word-overlap between contexts and adversarial negative responses; both datasets observe a score of 0.06. This low score indicates a is characteristic of adversarial negative responses. Furthermore, an examination of the adversarial negative responses within the aforementioned 1,000 samples confirmed their alignment with the predefined criteria for adversarial negative responses in the prompt. This comprehensive analysis substantiates the high quality of the newly generated positive and adversarial negative responses aimed at augmenting existing datasets.

\subsection{Baselines}
We select a range of widely used baseline metrics. The word-ovelap and embedding-based metrics includs BLEU~\cite{papineni-etal-2002-bleu}, ROUGE~\cite{lin-2004-rouge}, METEOR~\cite{banerjee-lavie-2005-meteor}, Embedding-Average~\cite{Wieting2016TowardsUP}, Vector-Extrema~\cite{forgues2014bootstrapping}, BERTScore~\cite{Zhang2020BERTScoreET}, Unieval \cite{zhong-etal-2022-towards}, and BARTScore \cite{yuan2021bartscore}.
For the LLM-based metrics, we select recently proposed baselines, including G-EVAL~\cite{liu-etal-2023-g} and another LLM-based metric proposed by \citet{Chiang2023CanLL} which we denote LLM-Chiang, as well as \texttt{Gemini}. 

\subsection{Evaluation Set}
As for the evaluation set, we select 600 context-response pairs from each of three open-domain dialogue datasets—namely
DailyDialog++, PersonaChat, and TopicalChat—resulting in a total of 1,800 samples. 
Then the evaluation set was evaluated by three human annotators, all of whom possess proficiency in English, to ensure an accurate assessment. 
They are instructed to thoroughly read each context-response pair and then rate them according to four distinct criteria including naturalness,
coherence, engagingness, groundedness, employing a 1-5 Likert scale for their assessments. 

The four criteria are consistent with ~\citet{zhong-etal-2022-towards} and are defined as follows: (1) \textbf{Naturalness}: The degree to which a response is naturally written; (2) \textbf{Coherence}: The extent to which the content of the output is well-structured, logical, and meaningful; 
(3) \textbf{Engagingness}: The degree to which the response is engaging; and (4) \textbf{Groundedness}: The extent to which a response is grounded in facts present in the context.

After obained four scores for each criteria, an aggregate score is calculated by averaging across all criteria to provide a comprehensive measure of each response's quality. Additionally, the Inner-Annotator Agreement (IAA) between three evaluators are evaluated through  Cohen's Kappa~\cite{doi:10.1177/001316446002000104}, which indicate a moderately strong agreement level (0.4-0.6) among annotators, with a value of 0.53, validating the reliability of the human evaluation.
% and the consistent standards applied by the annotators in assessing the dataset samples.

We employ \texttt{Gemini}, \texttt{GPT-3.5-turbo-1106} and \texttt{GPT-4-1106}  for all experiments. For SLM, DistilBERT~\cite{sanh2019distilbert} is utilised.

\subsection{Experimental Results} 

\subsubsection{Dialogue Classification Task}
We conduct dialogue classification on the on DailyDialog++ dataset as it contains the human-annotated labels for each context-response pair.
We calculate the classification accuracy using SLM and LLMs (i.e., \texttt{GPT-3.5} and \texttt{GPT-4}). The classification prompt for the LLMs could be found in ~\ref{prompt:class}. We also use SLM that trained on the identical dataset for the ablation study. 
% The final result is in
% \autoref{tab:classification}.

\begin{table}[ht]
\small
\resizebox{0.99\linewidth}{!}{
\begin{tabular}{l|c|c|c}
\toprule
\textbf{Model} & \multicolumn{1}{l|}{\textbf{Positive}} & \multicolumn{1}{l|}{\textbf{Negative}} & \multicolumn{1}{l}{\textbf{Overall}} \\
\midrule
\textbf{\texttt{GPT-3.5}}& 59.36 & 91.01 & 75.18\\
\textbf{\texttt{GPT-4}} & 80.43 & \textbf{97.40} & 88.91 \\
\textbf{\texttt{Gemini}} & 80.62 & 95.13 & 87.86 \\
\textbf{SLM (Dis)} & 79.35  & 90.03 & 84.00  \\ 
\textbf{SLM (Prob)} & 83.25 & 93.11& 88.19 \\ 
\textbf{SLM (Prob\& Dis)} & \textbf{91.83}& 90.28 & \textbf{91.05} \\
\bottomrule
\end{tabular}
}
\caption{Classification accuracy for only the Positive and adversarial negative responses, and the overall accuracy. Dis and Prob refers to the distance measures and classification probability, respectively.}
\label{tab:classification}
\end{table}

As Table \ref{tab:classification} shows, SLM employs both probability and distance measures, achieves superior performance with an accuracy of 91.05\%, which exceeds the 88.91\% accuracy achieved by \texttt{GPT-4}. 
Experimental results show that the SLM  particularly excels in the classification of positive responses, achieving the highest accuracy of 91.83\%. 
Conversely, \texttt{GPT-4} exhibits the strongest performance in classifying adversarial negative responses,  achieving 97.40\% accuracy. The performance of \texttt{Gemini} surpasses that of \texttt{GPT-3.5}, yet remains inferior to that of \texttt{GPT-4}. These results indicate that while LLMs, such as \texttt{GPT-4}, possess robust capabilities in recognising adversarial negative responses, they still struggle with the one-to-many problem, especially when dealing with semantically diverse positive responses. 
On the other hand, SLM with the integration of distance and probability metrics shows exceptional performance in recognising open-domain  positive responses. Based on these insights, we propose a hybrid approach that combines the SLM, a task-specific model, with LLMs for comprehensive dialogue evaluation. 
This approach is designed to leverage the distinct strengths of both types of models to enhance the overall performance in open-domain dialogevaluations.

\begin{table*}[htb]
\centering
\begin{threeparttable}[b]

\resizebox{0.99\linewidth}{!}{
\begin{tabular}{lcc|cc|cc}
             \toprule
& \multicolumn{2}{c|}{DailyDialog++} & \multicolumn{2}{c|}{PersonaChat} & \multicolumn{2}{c}{TopicalChat}\\ \midrule
Metrics & Pearson's $\rho$ & Spearman's $\tau$ & Pearson's $\rho$ & Spearman's $\tau$ & Pearson's $\rho$ & Spearman's $\tau$ \\ \midrule
%dd pc

BLEU-1 & 0.303 (<0.0001) & 0.242 (<0.0001) & 0.331 (<0.0001) & 0.276 (<0.0001) & 0.352 (<0.0001) & 0.330 (<0.0001)
        % 0.0280 (0.8025) & -0.1378 (0.2169) & 0.0567 (0.6770) & 0.0282 (0.8362) 
        \\
BLEU-2 & 0.317 (<0.0001) & 0.270 (<0.0001) & 0.311 (<0.0001) & 0.261 (<0.0001) & 0.279 (<0.0001) & 0.252 (<0.0001)
        % 0.0129 (0.9081) & -0.1320 (0.237) & -0.0235 (0.8621) & -0.0678 (0.6192) 
        \\
BLEU-3 & 0.256 (<0.0001) & 0.251 (<0.0001) & 0.251 (<0.0001) & 0.239 (<0.0001) & 0.179 (<0.0001) & 0.189 (<0.0001)
        % -0.0378 (0.7362) & -0.1508 (0.1761) & -0.0239 (0.8602) & -0.0539 (0.6927) 
        \\
BLEU-4 & 0.234 (<0.0001) & 0.247 (<0.0001) & 0.214 (<0.0001) & 0.217 (<0.0001) & 0.247 (<0.0001) & 0.154 (<0.0001)
        % 0.0134 (0.9047) & -0.0914 (0.4139) & -0.0896 (0.5101) & -0.0906 (0.5060) 
        \\ \midrule

ROUGE-1 & 0.303 (<0.0001) & 0.270 (<0.0001) & 0.304 (<0.0001) & 0.250 (<0.0001) & 0.374 (<0.0001) & 0.356 (<0.0001)
%       0.0623 (0.5784) & -0.1415 (0.2048) & 0.0460 (0.7360) & 0.1224 (0.3682) 
\\
ROUGE-2 & 0.178 (<0.0001) & 0.133 (<0.0001) & 0.142 (<0.0001) & 0.126 (<0.0001) & 0.173 (<0.0001) & 0.149 (<0.0001)
        % -0.0248 (0.8249) & -0.1538 (0.1677) & 0.0408 (0.7654) & 0.0548 (0.6880) 
        \\
ROUGE-L & 0.305 (<0.0001) & 0.271 (<0.0001) & 0.294 (<0.0001) & 0.244 (<0.0001) & 0.279 (<0.0001) & 0.259 (<0.0001)
        % 0.0473 (0.6728) & -0.1432 (0.1992) & 0.0251 (0.8534) & 0.1177 (0.3871) 
        \\ \midrule

METEOR & 0.196 (<0.0001) & 0.142 (<0.0001) & 0.250 (<0.0001) & 0.200 (<0.0001) & 0.293 (<0.0001) & 0.278 (<0.0001)
        % 0.1184 (0.2895) & 0.1630 (0.1435) & -0.0826 (0.5448) & -0.0800 (0.5576) 
        \\ \midrule
Embedding 
% & & & &
        \\
Extrema & 0.382 (<0.0001) & 0.367 (<0.0001) & 0.330 (<0.0001) & 0.299 (<0.0001) & 0.389 (<0.0001) & 0.244 (<0.0001)
        % 0.2617 (0.0174) & -0.0238 (0.8313) & -0.3019 (0.0231) & 0.1726 (0.2038)
    \\
Greedy & 0.322 (<0.0001) & 0.272 (<0.0001) & 0.328 (<0.0001) & 0.290 (<0.0001) & 0.272 (<0.0001) & 0.248 (<0.0001)
% 0.0850 (0.4433) & 0.1631 (0.1435) & -0.1070 (0.4320) & 0.013 (0.9231)
    \\
Average & 0.186 (<0.0001) & 0.208 (<0.0001) & 0.239 (<0.0001) & 0.249 (<0.0001) & 0.254 (<0.0001) & 0.261 (<0.0001)
        % -0.0070 (0.9433) & 0.0435 (0.6972) & -0.0461 (0.7350) & 0.0212 (0.8762)
    \\ \midrule
BERTScore & 0.419 (<0.0001) & 0.381 (<0.0001) & 0.465 (<0.0001) & 0.412 (<0.0001) & 0.525 (<0.0001) & 0.485 (<0.0001) \\
% 0.1968 (0.0701) & -0.1239 (0.2672) & 0.0251 (0.8531) & 0.0392 (0.7732) \\

BARTSCORE & 0.305 (<0.0001) & 0.264 (<0.0001) & 0.466 (<0.0001) & 0.442 (<0.0001) & 0.470 (<0.0001) & 0.427 (<0.0001)
% 0.1240 (0.2668) & 0.1482 (0.1837) & 0.0022 (0.9871) & -0.1006 (0.4603) 
\\
Unieval         & 0.103 (<0.0001) & 0.069 (<0.0001) & 0.192 (<0.0001) & 0.131 (<0.0001) & 0.357 (<0.0001) & 0.290 (<0.0001)             \\

\midrule

G-EVAL (\texttt{GPT-3.5}) & 0.447 (<0.0001) & 0.414 (<0.0001) & 0.516 (<0.0001) & 0.562 (<0.0001) & 0.642 (<0.0001) & 0.668 (<0.0001) \\

G-EVAL (\texttt{GPT-4}) & 0.641 (<0.0001) & 0.570 (<0.0001) & 0.597 (<0.0001) & 0.663 (<0.0001) & 0.696 (<0.0001) & 0.729 (<0.0001) \\

LLM-Chiang (\texttt{GPT-3.5}) & 0.632 (<0.0001) & 0.583 (<0.0001) & 0.599 (<0.0001) & 0.670 (<0.0001) & 0.667 (<0.0001) & 0.661 (<0.0001) \\

LLM-Chiang (\texttt{GPT-4}) & 0.723 (<0.0001) & 0.701 (<0.0001) & 0.701(<0.0001) & 0.675 (<0.0001) & 0.678 (<0.0001) & 0.709 (<0.0001) \\
\midrule

Ours (SLM-Prob-only) & 0.677 (<0.0001) & 0.632 (<0.0001) & 0.638 (<0.0001) & 0.692 (<0.0001) & 0.691 (<0.0001) & 0.689 (<0.0001) \\

Ours (SLM-Dis-only) & 0.652 (<0.0001) & 0.600 (<0.0001) & 0.668 (<0.0001) & 0.692 (<0.0001) & 0.642 (<0.0001) & 0.673 (<0.0001) \\

Ours (SLM-Dis and Prob) & 0.728 (<0.0001) & 0.666 (<0.0001) & 0.711 (<0.0001) & 0.707 (<0.0001) & 0.741 (<0.0001) & 0.727 (<0.0001) \\

Ours (LLMs-only (\texttt{GPT-3.5})) & 0.518 (<0.0001) & 0.472 (<0.0001) & 0.506 (<0.0001) & 0.617 (<0.0001) & 0.590 (<0.0001) & 0.669 (<0.0001) \\

SLIDE (\texttt{GPT-3.5})) & 0.727 (<0.0001) & 0.674 (<0.0001) & 0.692 (<0.0001) & 0.695 (<0.0001) & 0.696 (<0.0001) & 0.718 (<0.0001) \\

Ours (LLMs-only (\texttt{Gemini})) & 0.660 (<0.0001) & 0.614 (<0.0001) & 0.719 (<0.0001) & 0.688 (<0.0001) & 0.701 (<0.0001) & 0.699 (<0.0001) \\

SLIDE (\texttt{Gemini}) & 0.760 (<0.0001) & 0.689 (<0.0001) & 0.701 (<0.0001) & 0.705 (<0.0001) & 0.713 (<0.0001) & 0.712 (<0.0001) \\

Ours (LLMs-only (\texttt{GPT-4})) & 0.709 (<0.0001) & 0.666 (<0.0001) & \textbf{0.737 (<0.0001)} & \textbf{0.729 (<0.0001)} & \textbf{0.745 (<0.0001)} & \textbf{0.736 (<0.0001)} \\

SLIDE (\texttt{GPT-4}) & \textbf{0.773 (<0.0001)} & \textbf{0.704 (<0.0001)} & 0.713 (<0.0001) & 0.709 (<0.0001) & 0.711 (<0.0001) & 0.722 (<0.0001) \\

\bottomrule

\end{tabular}
}
\end{threeparttable}
\caption{Pearson and Spearman correlations with human judgements. Figures in parentheses are p-values. Dis and Prob refers to the distance measures and classification probability, respectively. 
}
\label{tab:similar_correlation}
\end{table*}

% Our ablation studies include: "Ours (SLM-Prob-only)" only rely on the probability value as the evaluation metric. "Ours (Distance-only)" employs only the normalised distance for assessment. "Ours (LLM-only (\texttt{GPT-3.5}))" signifies the sole use of \texttt{GPT-3.5}, whereas "Ours (LLM-only (\texttt{GPT-4}))" indicates the exclusive use of \texttt{GPT-4}, without incorporating SLM. SLIDE represents our integrated model that includes SLM, leveraging both probability and distance, along with LLM.

\subsubsection{Dialogue Evaluation}
As shown in Table \ref{tab:similar_correlation}, we investigate the correlation between automatic evaluation metrics and human judgments across three datasets. The word-overlap metrics, including BLEU, ROUGE, and METEOR, exhibit a weak positive correlation with human scores (<0.4), among which ROUGE-2 showing the least correlation (<0.2). Similarly, embedding-based metrics, such as Embedding and BARTScore,  demonstrate a weak correlation with human evaluations, ranging from 0.2 to 0.4. Notably, BERTScore shows the strongest correlation to human judgments among both word-overlap and embedding based traditional metrics. On the other hand, Unieval achieves the suboptimal results (<0.3) on the DailyDialog++ and PersonaChat datasets .

As for LLM-Based metrics, there is a notable improvement in correlation with human scores compared with traditional metrics, most show correlations greater than 0.6. Specifically, Ours (LLMs-only(\texttt{GPT-4}) exhibits superior performance compared with G-EVAL and LLM-Chiang and achieves the highest correlation in both the PersonaChat (0.737 for Pearson and 0.729 for Spearman) and TopicalChat datasets (0.745 for Pearson and 0.736 for Spearman). 
Moreover, \
% texttt{GPT-4} outperforms \texttt{GPT-3.5}, albeit at a higher financial cost. 
the variance in performance between G-EVAL and LLM-Chiang indicates the significant impact of model selection (i.e., \texttt{GPT-4} or \texttt{GPT-3.5}) and prompt design on LLM-based metric effectiveness .

As for the SLM, Our SLM-Dis and Prob approach that incorporate both distance and probability measures, revealing a strong correlation with human evaluations (>0.6). 
% These metrics, which blend distance and probability, either equal to or exceed the performance of LLM-based metrics, with minimal disparity observed. 
As for the SLIDE method, it combines an SLM with LLMs. SLIDE (\texttt{GPT-4} ) achieves the strongest correlation with human ratings on the DailyDialog++ dataset (0.773 for Pearson and 0.704 for Spearman).
% , along with commendable results on other LLM-generated datasets(>0.7). 
Furthermore, our LLM metrics, characterised by simpler prompt designs compared to those used in G-EVAL and LLM-Chiang, do not individually surpass these benchmarks. 
However, the amalgamation of LLM and SLM techniques results in a significant enhancement in correlation strength, surpassing both G-EVAL and LLM-Chiang, particularly when employing \texttt{GPT-3.5}. The same trend is also observed when utilising \texttt{Gemini}. These two distinct models could demonstrate the effectiveness of integrating LLM and SLM techniques.

The datasets generated by \texttt{GPT-4}, namely PersonaChat test set and Topical-Chat test set, exhibit the strongest correlations with the \texttt{GPT-4}-based metrics, including our proposed method. 
It indicates the inherent advantage of using model-specific metrics for evaluating datasets generated by the same model, thereby leading to a more accurate and relevant assessment of these two datasets.

To validate the effectiveness of distance metric used in SLM and SLIDE, we conduct several ablation studies. 
It can be observed that Ours (SLM-Dis-Prob) gives better performance than
the model variant with the Dis (SLM-Dis-only) or Prob (SLM-Prob-only) component only.
This suggests that SLM with distance metric is more effective in
evaluating open-domain dialogues.

% Ours (SLM-Prob-only)" only rely on the probability value as the evaluation metric. "Ours (Distance-only)" employs only the normalized distance for assessment. "Ours (LLM-only (\texttt{GPT-3.5}))" signifies the sole use of \texttt{GPT-3.5}, whereas "Ours (LLM-only (\texttt{GPT-4}))" indicates the exclusive use of \texttt{GPT-4}, without incorporating SLM. SLIDE represents our integrated model that includes SLM, leveraging both probability and distance, along with LLM.
% \section*{Acknowledgements}
\begin{figure*}[h]
\centering
    % \subfigure[]{
    % \begin{subfigure}[]{0.4\textwidth}
	\includegraphics[scale=0.45]{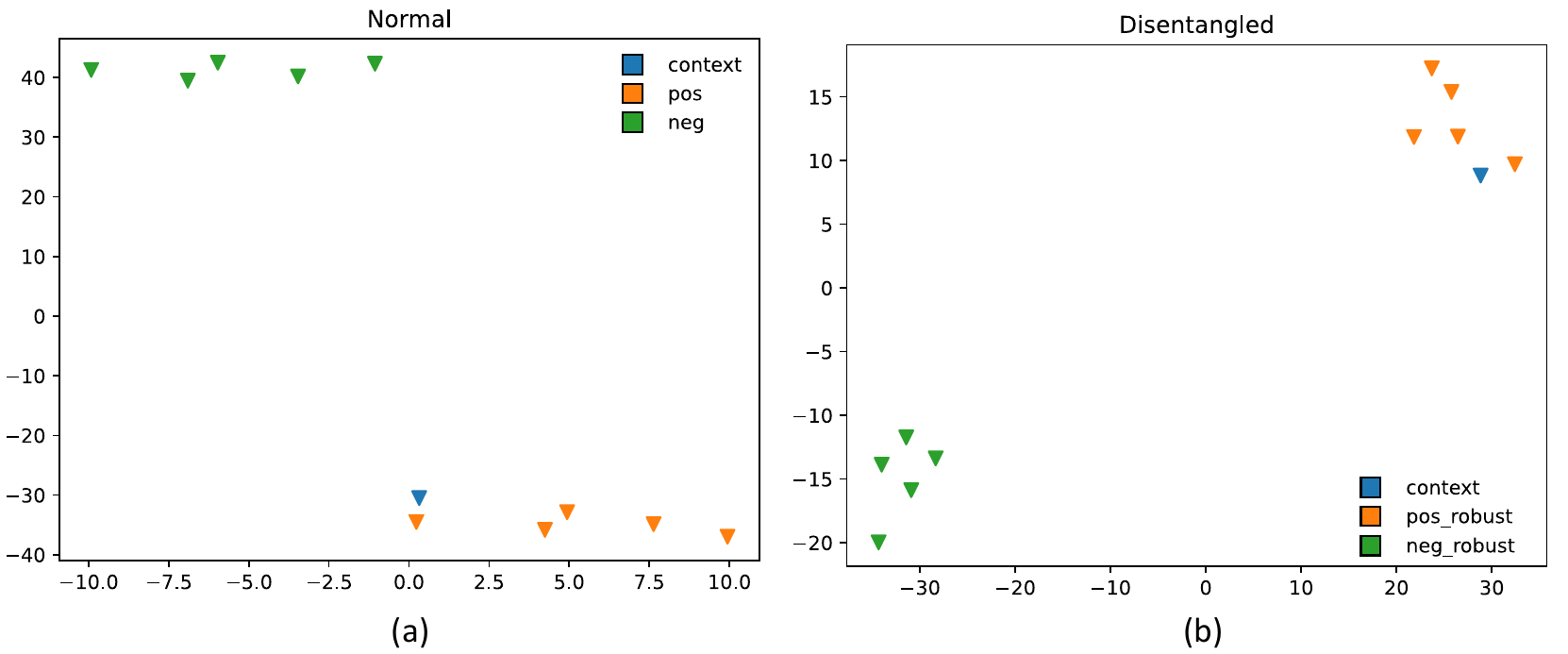}
% 	\caption{}
% 	\end{subfigure}
% % 	\subfigure[]{
%     \begin{subfigure}[]{0.4\textwidth}
% 	\includegraphics[scale=0.4]{pic/fig1_2.pdf}
% 	\caption{}
% 	\end{subfigure}%
% 	\hfill
% 	\subfigure[]{

% 	\subfigure[]{
\caption{T-SNE visualisation of the sentence representation of context and responses. The left panel, labeled \textit{Normal}, illustrates the vectors prior to disentanglement, whereas the right panel, labeled \textit{Disentangled}, displays the post-disentanglement outcomes. This demonstrates the convergence of negative responses towards the context following disentanglement.}
\label{fig:label}
\end{figure*}
% \begin{figure*}[ht]
% \small
% \centering 
%     % \subfigure[]{
%     \begin{subfigure}[]{0.4\textwidth}
% 	\includegraphics[scale=0.4]{pic/fig1_1.pdf}
% 	\caption{}
% 	\end{subfigure}
% % 	\subfigure[]{
%     \begin{subfigure}[]{0.4\textwidth}
% 	\includegraphics[scale=0.4]{pic/fig1_2.pdf}
% 	\caption{}
% 	\end{subfigure}%
% % 	\hfill
% % 	\subfigure[]{

% 	\begin{subfigure}[]{0.4\textwidth}
% 	\includegraphics[scale=0.4]{pic/fig2_1.pdf}
%     \caption{}
%     \end{subfigure}%
% % 	\subfigure[]{
%     \begin{subfigure}[]{0.4\textwidth}
% 	\includegraphics[scale=0.4]{pic/fig2_2.pdf}
% 	\caption{}
% 	\end{subfigure}%
% \caption{T-SNE visualisation of the sentence representation of context and responses for 2 examples. Each example includes a context, five positive responses, and five adversarial negative responses. The left represents the vector prior to disentanglement which is titled "Normal", whilst the right is after disentanglement which is titled "Disentangled". These figures demonstrate the positive responses become nearer to context after disentangling.}
% \label{fig:label}
% \end{figure*}
\subsection{Example Visualisation}
% In order to display the effectiveness of disentanglement, we select some examples from the DailyDialog++ dataset and use T-SNE to visualise them in \autoref{fig:label}. 
To demonstrate the effectiveness of disentanglement, we selected a subset of examples from the DailyDialog++ dataset and employed T-SNE for visualisation, as depicted in Figure~\ref{fig:label}.
Each example includes a context, five positive responses, and five adversarial negative responses.
% The left plot is the representation vector before disentanglement, whilst the right plot is the representation after. 
The vector representations prior to disentanglement are illustrated in the left plot, whereas those post-disentanglement are shown on the right. 
Post-disentanglement visualisations reveal a more distinct clustering of responses in relation to their contexts. By utilising these refined representations to train a classifier, we observed an improvement in classification accuracy on the test set of DailyDialog++ dataset, from 87\% to 88.19\%. Additional examples are available in Appendix~\ref{app:vic}.

\begin{table}[ht]
\scriptsize
\begin{tabular}{cccl}
\toprule[1pt]
\multicolumn{1}{l}{\textbf{Context:}}   & \multicolumn{3}{l}{\begin{tabular}{p{4.5cm}}FS: We have another traditional holiday-the Dragon Boat Festival. 
\\SS: When is it? 
\\FS: It falls on the fifth day of the fifth lunar month.\end{tabular}}\\
\midrule[1pt]
\multicolumn{1}{l}{\textbf{Reference:}} & \multicolumn{3}{l}{\begin{tabular}{p{4.5cm}}Oh! That is great.\end{tabular}} \\ 
\midrule[1pt]
\multicolumn{1}{l}{\textbf{Response:}} & \multicolumn{3}{l}{\begin{tabular}{p{4.5cm}} I did not hear about that.\end{tabular}} \\ 
\midrule[1pt]
 Human & G-EVAL (\texttt{GPT-3.5}) &  G-EVAL (\texttt{GPT-4})\\ 
4.70 & 2.50 & \multicolumn{1}{c}{2.25} \\ 
\hline
 Chiang (\texttt{GPT-3.5}) & Chiang (\texttt{GPT-4}) & \multicolumn{1}{c}{SLM} \\
2.25 & 2.75 & \multicolumn{1}{c}{4.60} \\ 
\hline
 Ours (w/o LLM) & SLIDE (\texttt{GPT-3.5}) & \multicolumn{1}{c}{SLIDE (\texttt{GPT-4})} \\
5.00 & 4.30 & \multicolumn{1}{c}{5.00} \\

\midrule[1pt]

\multicolumn{1}{l}{\textbf{Context:}} & \multicolumn{3}{l}{\begin{tabular}{p{4.5cm}}FS: We have been over this a hundred times! We are not getting a pet! 
\\SS: Why not? Come on! Just a cute little puppy or a kitty! 
\\FS: Who is going to look after a dog or a cat?\end{tabular}} \\ 
\midrule[1pt]
\multicolumn{1}{l}{\textbf{Reference:}} & \multicolumn{3}{l}{\begin{tabular}{p{4.5cm}}We both will look after it.\end{tabular}} \\ 
\midrule[1pt]
\multicolumn{1}{l}{\textbf{Response:}} & \multicolumn{3}{l}{\begin{tabular}{p{4.5cm}}Will you look after me once I get old?\end{tabular}} \\ 
\midrule[1pt]
Human & G-EVAL (\texttt{GPT-3.5}) & G-EVAL (\texttt{GPT-4})\\ 
2.00 & 1.25 & 2.25 \\ 
\hline
Chiang (\texttt{GPT-3.5}) & Chiang (\texttt{GPT-4}) & SLM \\
4.00 & 1.25 & \multicolumn{1}{c}{1.00} \\ 
\hline
Ours(w/o LLM) & SLIDE (\texttt{GPT-3.5}) & \multicolumn{1}{c}{SLIDE (\texttt{GPT-4})} \\
2.00 & 2.79 & \multicolumn{1}{c}{2.40} \\ 

\bottomrule[1pt]
\end{tabular}
\caption{Samples from the DailyDialog++ dataset. "FS" is the First Speaker and "SS" is the Second Speaker}
\label{tab:case study1}
\end{table}

\subsection{Case Study}
We conduct qualitative analyses through case studies, which are shown in ~\autoref{tab:case study1}. 
Each case shows the conversational context as well as the corresponding gold-standard reference and thegenerated responses. 
We compare our evaluation metric with nine different baselines. 
To simplify the comparison, we normalise all scores to a range of 1-5 Likert scale. 
Note that the normalisation is only applied to the case study, and is not implemented in our main experiments.
In the first case, the response candidate is positive, whereas the response is an adversarial negative response in the second case.
Our SLIDE metric correlate well with the human ratings. 
For example, in the first scenario, only our models rate a positive score (>3), which is consistent with the human score. 
% However, while other LLM-based metrics give a negative score (<3). 
More examples could be found in Appendix~\ref{app:case}.

\section{Conclusion}
In this paper, we introduce a novel automatic evaluation metric (SLIDE) that integrates SLM with LLMs for open-domain dialogue evaluation. 
Our approach involves initially training a SLM through iterative  contrastive learning stages, followed by leveraging the combine strengths of SLM and LLMs to create a superior evaluation metric. 
This metric exhibits enhanced correlation with human judgments in comparison to those derived exclusively from LLMs.
% compared to those based solely on LLMs in regard to correlations with human judgements. 
Furthermore, experimental results reveal that LLMs-based metric struggles with classifying and evaluating open-domain dialogues, which is the one-to-many nature. Therefore, we design a novel method for merging scores from SLM and LLM to enhance dialogue evaluation. 
Experimental results show that our SLIDE model outperforms a wide range of baseline methods in terms of both Pearson and Spearman correlations 
with human judgements on three open-domain dialogue datasets, and deals well with the one-to-many issue in open-domain dialogue evaluation.
% Experimental evidence demonstrates that our SLIDE sets a new benchmark in performance on human-annotated multi-reference datasets and shows a strong correlation with human judgments across three datasets.

\section*{Ethics Statement}

In this paper, we introduce SLIDE, a novel automatic evaluation metric that integrates SLM and LLM for assessing open-domain dialogue systems. The advantage of SLIDE lies in its utilisation of both SLM and LLM, enhancing the evaluation of open-domain dialogues. However, a potential downside is that SLIDE might award high scores to responses that are inappropriate or offensive under certain conditions. Therefore, it is crucial to carefully review the content of training datasets prior to training SLIDE to mitigate this issue.

\section*{Limitations}

Although our proposed method performs well in evaluating the open-domain dialogue systems, it also has some limitations. Primarily. Although our model first combines SLMs and LLMs for automatic evaluation metrics, it is a simple combination according to its characteristics. In the future, we will conduct a deeper combination of LLM and SLM to make a better evaluation on the open-domain dialogue system. In addition,  because the PersonaChat dataset and TopicalChat dataset are augmented by LLM, not the real human-annotated dataset, our model does not perform the best on these two datasets. We need to further employ the difference between LLM-generated datasets and human-annotated dataset, and further analyse the effectiveness of our proposed model.

\section*{Acknowledgments}
This study was partially supported by the National Science Foundation (IIS 2045848 and 2319450). Part of the work used Bridges-2 at Pittsburgh Supercomputing Center through bridges2 \cite{brown2021bridges} from the Advanced Cyberinfrastructure Coordination Ecosystem: Services \& Support (ACCESS) program, which is supported by NSF grants \#2138259, \#2138286, \#2138307, \#2137603, and \#2138296.
\bibliography{custom}

\appendix
\section{Appendices}

\subsection{Prompt for evaluating the engagingness of the response }
\label{prompt:eva}

\textit{Engagingness\\
Please rate the dialogue response.\\
The goal of this task is to rate dialogue response.\\
Note: Please take the time to fully read and understand the  dialogue response. We will reject submissions from workers that are clearly spamming the task.\\
How engaging is the text of the dialogue response? (on a scale of 0-1, with 0 being the lowest)\\\\
Example:\\
Conversation History:\\
Is there something wrong? \\
I enjoy having your daughter in my class. \\
I'm glad to hear it.
\\\\
Dialogue Response:\\
I enjoy listening jazz music in my free time.
\\\\
Evaluation Form (scores ONLY): \\
Engagingness: 1
\\\\
Input:\\
Conversation History:\\
${}$\\
\\
Response:\\
${}$\\\\
Evaluation Form (scores ONLY):\\
Engagingness: \\
}
% You are an evaluator. Given a conversational context including 2 speakers (annotated as FS for First Speaker and SS for Second Speaker) and a response, your task is to score the following dialogue response on a continuous scale from 0.0 to 1.0. You do not need to give an explanation, so you only need to give the score.''}
\subsection{Prompt for generating response}
\label{prompt:generation}
\textit{You are a conversational dialogue generator. \\
Given a conversation context, , which includes 2 speakers[annotated as FS(FirstSpeaker) and SS(SecondSpeaker)], \
and a response.\\
Your task is to generate five diverse positive response and five adversarial negative response respectively. 
\\\\
Positive Response\\
Positive response is valid for the conversation context.\\\\
Adversarial Negative Response\\
Adversarial negative responses have a significant word overlap with the conversation context but are still irrelevant response, which may not have any relation to the context.
You need to choose some words (do not include stopwords such as "I", "you", "are", etc.) from the conversation context and use them directly or indirectly while writing the adversarial negative responses. 
Indirect usage here refers to using words closely related to the context words. \\
For example,using synonyms,antonyms, homonyms, subwords, or other words that are known to frequently co-occur with the words in the context (e.g., the words  "flexibility" and "injuries" co-occur with "acrobatics").
\\\\
The following are five examples of a conversation context and response, and the corresponding prediction. \\\\
Example\\
Context: \\
FS:Is there something wrong?\\
SSI enjoy having your daughter in my class.\\
FS:I'm glad to hear it.\\
Positive response: 
She is so brilliant.\\
Her behavior is good in the class.\\
I would love to hear that she knows every rules and regulation.\\
I was shocked to know that she is your daughter.\\
She answers all my questions.\\\\
Adversarial Negative Responses:\\
I enjoy listening jazz music in my free time.\\
I need pin drop silence in the class. \\
If I hear someone talking they will be sent out of the class.\\
I am glad you enjoyed the magic show organised by our team. \\
I think there was something wrong with the CCTV camera installed in the class.\\
This is the wrong method to solve the problem. Please be attentive in the class.}
% \textit{``You are a dialogue generators. Given the conversation contexts, which includes 2 speakers[annotated as FS(First Speaker) and SS(Second Speaker)], you need to generate 5 positive responses and 5 adversarial negative responses according to this conversation.These generated responses should semantically different. For positive responses, it means the response is valid for this conversation. For adversarial response, you need to choose some words from the conversation context and use them directly or indirectly while writing the responses. Indirect usage here refers to using words closely related to the context words. For example, using synonyms,antonyms, homonyms, subwords, or other words that are known to frequently co-occur with the words in the context (e.g., the words ‘‘flexibility’’ and ‘‘injuries’’ co-occur with ‘‘acrobatics’’). You only need to generate the response.'' }

\subsection{Prompt for classifying response}
\label{prompt:class}
\textit{
You are a classifier. Given a conversation context, which includes 2 speakers[annotated as FS(FirstSpeaker) and SS(SecondSpeaker)], \
and a response. Your task is to classify this response whether is positive or negative. \
\\\\
Positive Response \\
Positive response is valid for the conversation context. 
\\\\
Adversarial Negative Response\\
Adversarial negative responses have a significant word overlap with the conversation context but are still irrelevant response, which may not have any relation to the context.
You need to choose some words (do not include stopwords such as "I", "you", "are", etc.) from the conversation context and use them directly or indirectly while writing the adversarial negative responses. 
Indirect usage here refers to using words closely related to the context words. \\
For example,using synonyms,antonyms, homonyms, subwords, or other words that are known to frequently co-occur with the words in the context (e.g., the words  "flexibility" and "injuries" co-occur with "acrobatics").
\\\\
Your output format is only the “Positive” or “Negative”.
\\\\
Example\\
The following are five examples of a conversation context and response, and the corresponding prediction. \
\\\\
Example 1:\\\\
Context: \\
FS: Is there something wrong?\\
SS: I enjoy having your daughter in my class.\\
FS: I'm glad to hear it.
\\\\
Response: \\
She is so brilliant.\\
Prediction: Positive
\\\\
Example 2:\\
Context: \\
FS: Is there something wrong?\\
SS: I enjoy having your daughter in my class.\\
FS: I'm glad to hear it.
\\\\
Response: \\
I enjoy having your daughter in my class.\\
Prediction: Positive
\\\\
Example 3:\\
Context: \\
FS: Is there something wrong?\\
SS: I enjoy having your daughter in my class.\\
FS: I'm glad to hear it.
\\\\
Response: \\
I'm glad to hear it.\\
Prediction: Positive
\\\\
Example 4:\\
Context: \\
FS: We have to pick up Conrad before the party.\\
SS: Alright, no problem.\\
FS: We're supposed to meet him at Cal's Bar at 10
\\\\
Response: \\
I pushed the problem aside; at present it was insoluble."\\
Prediction: Negative
\\\\
Example 5:\\
Context: \\
FS: Is there something wrong?\\
SS: I enjoy having your daughter in my class.\\
FS: I'm glad to hear it.
\\\\
Response: \\
I think there was something wrong with the CCTV camera installed in the class.\\
Prediction: Negative\\
}

\subsection{Case Study}
\label{app:case}

We display some other samples from DailyDialog++ dataset. From Table \ref{tab:case study2}, We can see that our SLIDE model have a close score with human score.

\begin{table}[ht]
\scriptsize
\begin{tabular}{cccl}
\toprule[1pt]
% \multicolumn{1}{l}{\textbf{Context:}}   & \multicolumn{3}{l}{\begin{tabular}{p{4.2cm}}FS: Of course . Let me tell you in some detail about our idea. You know the popular Hello Kitty products. \\SS: Yes, of course. \\FS: Well, the products in themselves are very simple.It is the logo that is successful.So, Hello Kitty is successful because of the logo, but the products are very simple.\end{tabular}}\\                                                                          \midrule[1pt]
% \multicolumn{1}{l}{\textbf{Reference:}} & \multicolumn{3}{l}{\begin{tabular}{p{4.2cm}}Well, the brand name plays an important role in business.\end{tabular}} \\ \midrule[1pt]
% \multicolumn{1}{l}{\textbf{Response:}} & \multicolumn{3}{l}{\begin{tabular}{p{4.2cm}} Yeah, I don't think so, the products must be good.\end{tabular}}                               \\ \midrule[1pt]
% Human & G-EVAL(\texttt{GPT-3.5}) &G-EVAL(\texttt{GPT-4})\\ 
% 4.00         & 2.25 & \multicolumn{1}{c}{2.00} \\ \hline                                                                                 
% Chiang (\texttt{GPT-3.5})  & Chiang (\texttt{GPT-4})   & \multicolumn{1}{c}{SLM}  \\
% 2.50      & 1.75    & \multicolumn{1}{c}{4.60} \\ \hline

% Ours (w/o LLM) & SLIDE (\texttt{GPT-3.5}) & \multicolumn{1}{c}{SLIDE (\texttt{GPT-4})} \\
% 5.00 & 4.40 & \multicolumn{1}{c}{4.00} \\

% \midrule[1pt]

\multicolumn{1}{l}{\textbf{Context:}}   & \multicolumn{3}{l}{\begin{tabular}{p{4.5cm}}FS: We have been over this a hundred times! We are not getting a pet!\\ SS: Why not? Come on! Just a cute little puppy or a kitty! \\FS: Who is going to look after a dog or a cat?\end{tabular}}                                                                                                                            \\ \midrule[1pt]
\multicolumn{1}{l}{\textbf{Reference:}} & \multicolumn{3}{l}{\begin{tabular}{p{4.5cm}}We both will look after it.\end{tabular}} \\ \midrule[1pt]
\multicolumn{1}{l}{\textbf{Response:}} & \multicolumn{3}{l}{\begin{tabular}{p{4.5cm}}Will you look after me once I get old?\end{tabular}}                                                                                                                     \\ \midrule[1pt]
Human & G-EVAL (\texttt{GPT-3.5}) &G-EVAL (\texttt{GPT-4})                                                                                    \\ %\hline
2.00          & 1.25 & \multicolumn{1}{c}{2.25} \\ \hline                                                                                 
Chiang (\texttt{GPT-3.5})  & Chiang (\texttt{GPT-4})   & \multicolumn{1}{c}{SLM}  \\
4.00      & 1.25    & \multicolumn{1}{c}{1.00} \\ \hline

Ours (w/o LLM) & SLIDE (\texttt{GPT-3.5}) & \multicolumn{1}{c}{SLIDE (\texttt{GPT-4})} \\
2.00 & 2.79 & \multicolumn{1}{c}{2.40} \\ 
\midrule[1pt]
\multicolumn{1}{l}{\textbf{Context:}}   & \multicolumn{3}{l}{\begin{tabular}{p{4.5cm}}FS: Why not? We're supposed to meet him there. \\SS: Why doesn't he meet us outside? \\FS: Why should he do that? It isn't illegal for us to go in.\end{tabular}}                                                                                                                           \\ \midrule[1pt]
\multicolumn{1}{l}{\textbf{Reference:}} & \multicolumn{3}{l}{\begin{tabular}{p{4.5cm}}I really don’t want to go to that place.\end{tabular}} \\ \midrule[1pt]
\multicolumn{1}{l}{\textbf{Response:}} & \multicolumn{3}{l}{\begin{tabular}{p{4.5cm}}Nothing wrong in going there, so let's go there and meet him.daughter.\end{tabular}}                                                                                                                     \\ \midrule[1pt]
Human & G-EVAL (\texttt{GPT-3.5}) &G-EVAL (\texttt{GPT-4})                                                                                    \\ %\hline
5.00          & 2.25 & \multicolumn{1}{c}{2.00} \\ \hline                                                                                 
Chiang (\texttt{GPT-3.5})  & Chiang (\texttt{GPT-4})   & \multicolumn{1}{c}{SLM}  \\
3.50      & 3.75    & \multicolumn{1}{c}{4.90} \\ \hline

Ours (w/o LLM) & SLIDE (\texttt{GPT-3.5}) & \multicolumn{1}{c}{SLIDE (\texttt{GPT-4})} \\
5.00 & 5.00 & \multicolumn{1}{c}{4.80} \\

\bottomrule[1pt]
\end{tabular}
\caption{Samples from the DailyDialog++ dataset. "FS" is the First Speaker and "SS" is the Second Speaker}
\label{tab:case study2}
\end{table}

\subsection{Example Visualisation}
\label{app:vic}

\begin{figure*}[ht]
\small
\centering 
    % \subfigure[]{
    \begin{subfigure}[]{0.45\textwidth}
	\includegraphics[scale=0.45]{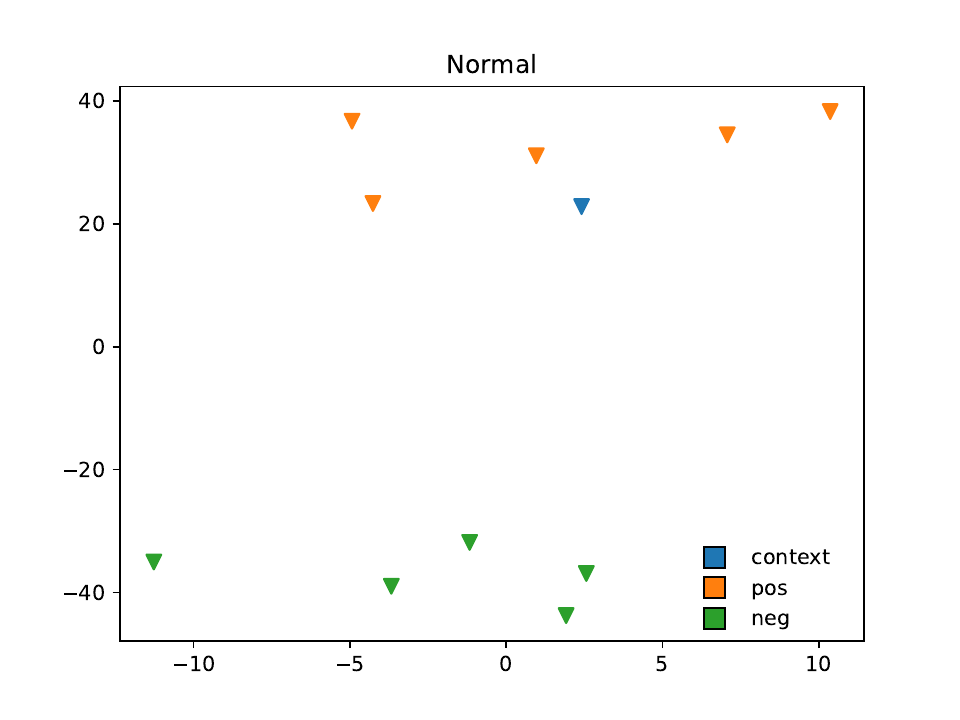}
	\caption{}
	\end{subfigure}
% 	\subfigure[]{
    \begin{subfigure}[]{0.45\textwidth}
	\includegraphics[scale=0.45]{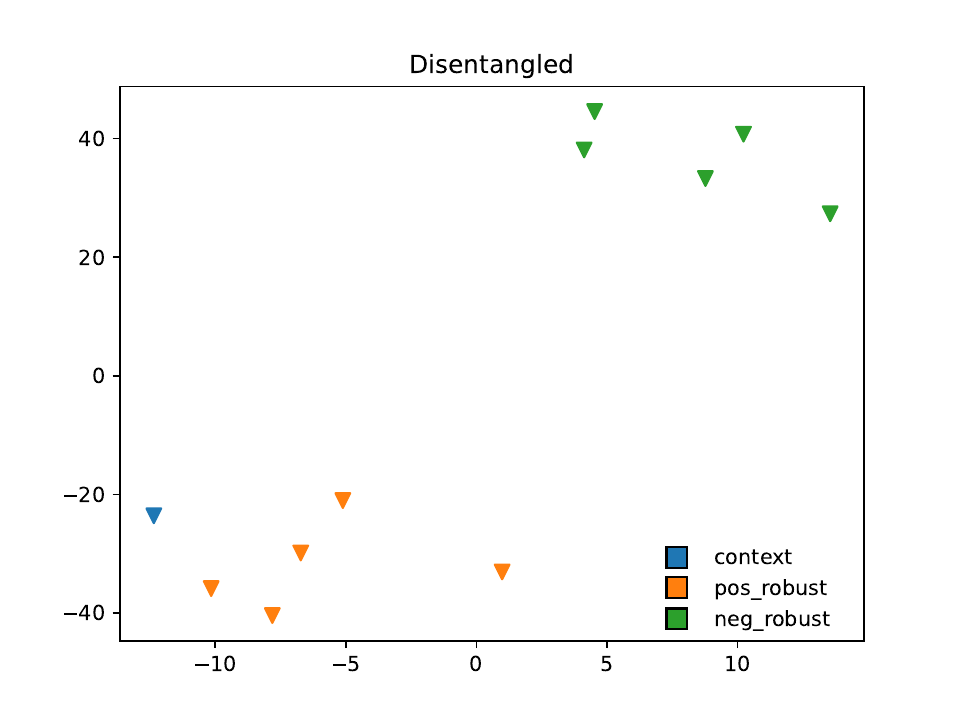}
	\caption{}
	\end{subfigure}%
% 	\hfill
% 	\subfigure[]{

	\begin{subfigure}[]{0.45\textwidth}
	\includegraphics[scale=0.45]{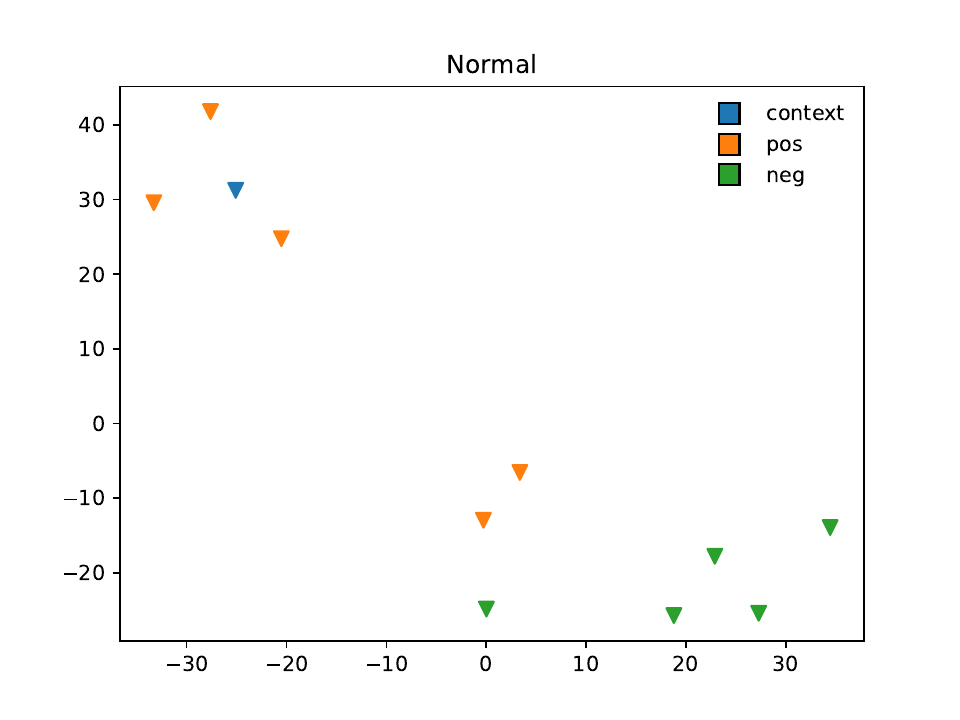}
    \caption{}
    \end{subfigure}%
% 	\subfigure[]{
    \begin{subfigure}[]{0.45\textwidth}
	\includegraphics[scale=0.45]{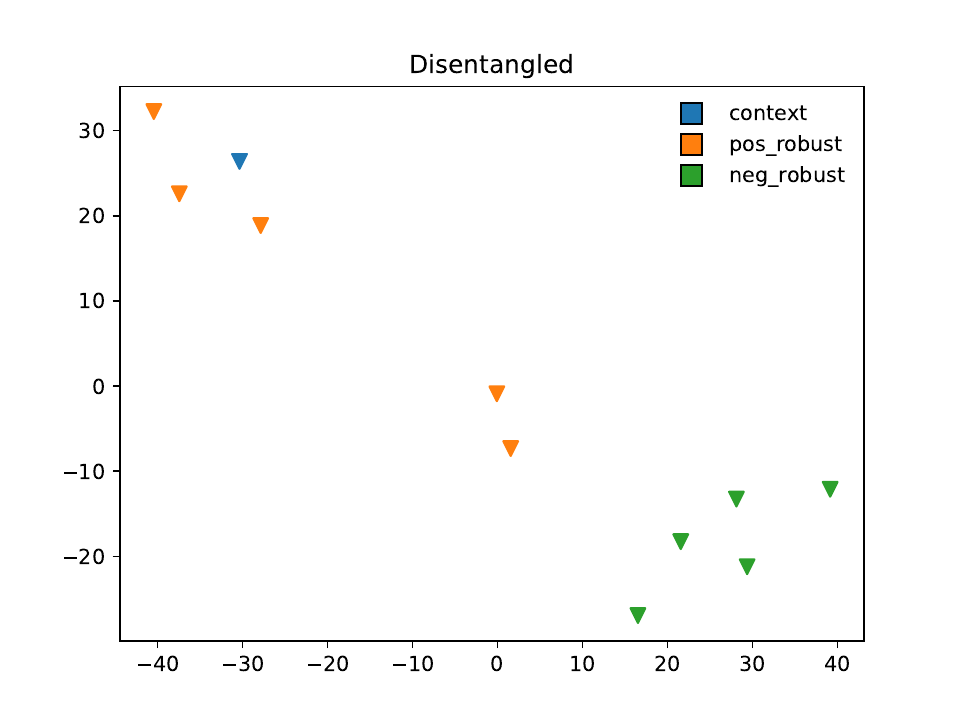}
	\caption{}
	\end{subfigure}%

    \begin{subfigure}[]{0.45\textwidth}
	\includegraphics[scale=0.45]{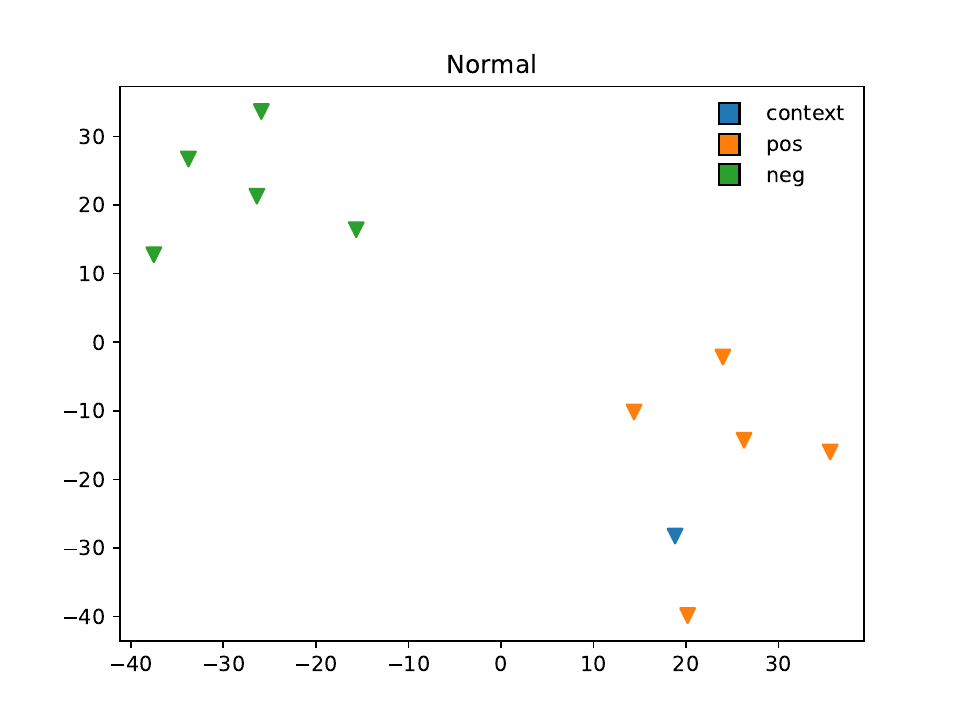}
    \caption{}
    \end{subfigure}%
% 	\subfigure[]{
    \begin{subfigure}[]{0.45\textwidth}
	\includegraphics[scale=0.45]{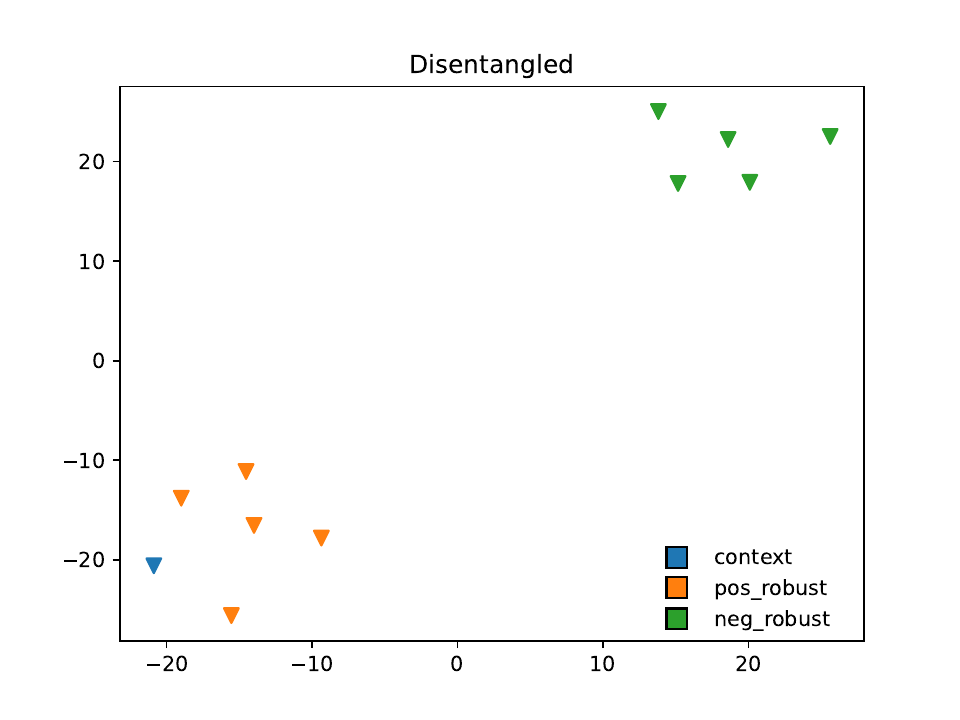}
	\caption{}
	\end{subfigure}%
 
     \begin{subfigure}[]{0.45\textwidth}
	\includegraphics[scale=0.45]{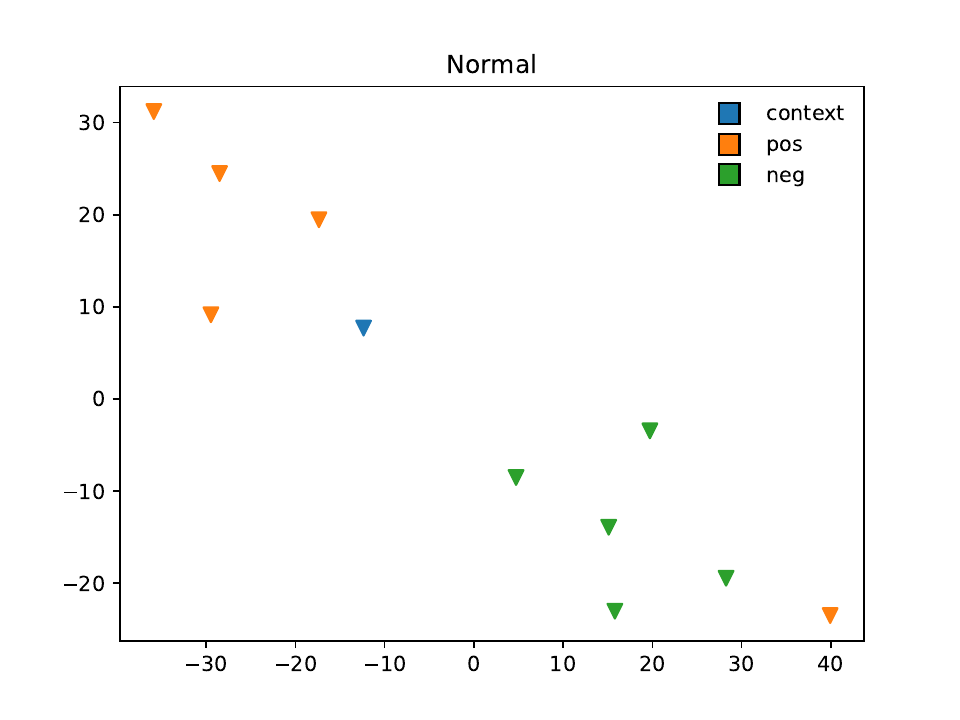}
    \caption{}
    \end{subfigure}%
% 	\subfigure[]{
    \begin{subfigure}[]{0.45\textwidth}
	\includegraphics[scale=0.45]{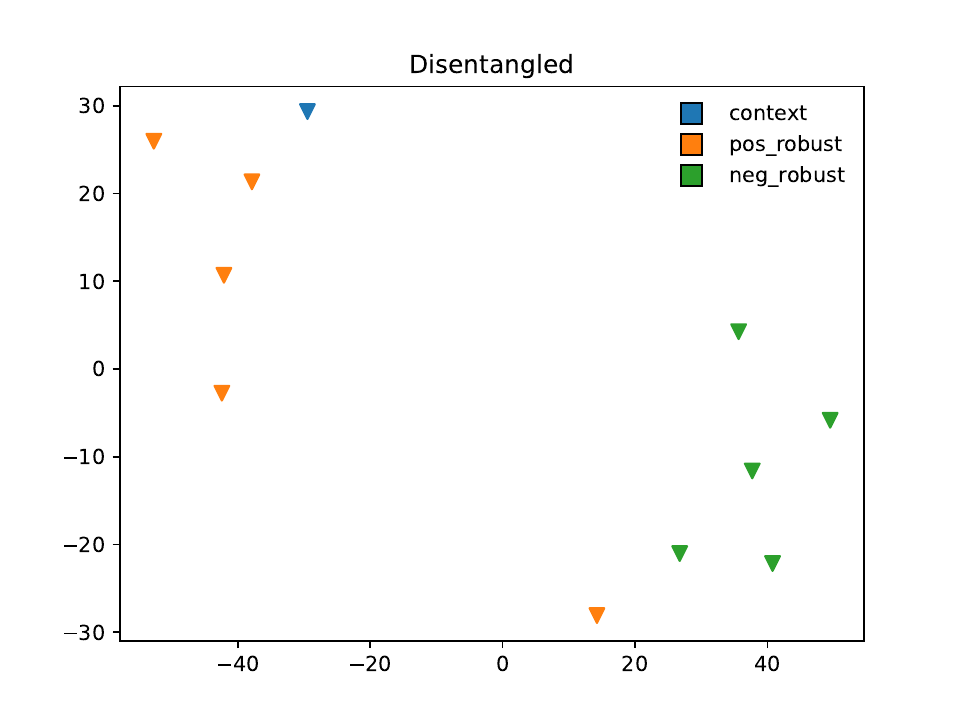}
	\caption{}
	\end{subfigure}%
\caption{T-SNE visualisation of the sentence representation of context and responses for some examples. Each example includes a context, five positive responses, and five adversarial negative responses. The left represents the vector prior to disentanglement which is titled "Normal", whilst the right is after disentanglement which is titled "Disentangled". These figures demonstrate the positive responses become nearer to context after disentangling.}
\label{fig:label1}
\end{figure*}

We display some other visualization example in this section. From Figure \ref{fig:label1}, We can see that the positive responses become closer to context after disentanglement.

\end{document}